\documentclass[letterpaper, 10 pt, conference]{ieeeconf}
\usepackage{PackageHeading}
\usepackage[absolute,overlay]{textpos}

\newcommand{\Sajad}[1]{\textcolor{black}{#1}}

\title{\Large \bf
Parallel Reinforcement Learning Simulation for Visual Quadrotor Navigation
}

\author{Jack Saunders$^{1}$, Sajad Saeedi$^{2}$, and Wenbin Li$^{1}$% <-this % stops a space
\thanks{This work is supported by the UKRI Centre for Doctoral Training in Accountable, Responsible \& Transparent AI (ART-AI), under UKRI grant number EP/S023437/1.}% <-this % stops a space
\thanks{$^{1}$Department of Computer Science,
        University of Bath, UK, \{\protect\url{js3442, wl281}\}\protect\url{@bath.ac.uk}}%
\thanks{$^{2}$Department of Mechanical and Industrial Engineering, Toronto Metropolitan University,
        Toronto, Canada, \{\protect\url{s.saeedi}\}\protect\url{@ryerson.ca}}%
}

\begin{document}
\maketitle
\thispagestyle{empty}
\pagestyle{empty}

\begin{abstract}
Reinforcement learning (RL) is an agent-based approach for teaching robots to navigate within the physical world.  Gathering data for RL is known to be a laborious task, and real-world experiments can be risky.  Simulators facilitate the collection of training data in a quicker and more cost-effective manner.  However, RL frequently requires a significant number of simulation steps for an agent to become skilful at simple tasks.  This is a prevalent issue within the field of RL-based visual quadrotor navigation where state dimensions are typically very large and dynamic models are complex.  Furthermore, rendering images and obtaining physical properties of the agent can be computationally expensive.  To solve this, we present a simulation framework, built on AirSim, which provides efficient parallel training.  Building on this framework, Ape-X is modified to incorporate decentralised training of AirSim environments to make use of numerous networked computers.  Through experiments we were able to achieve a reduction in training time from 3.9 hours to 11 minutes using the aforementioned framework and a total of 74 agents and two networked computers.  Further details including a github repo and videos about our project, PRL4AirSim, can be found at \url{https://sites.google.com/view/prl4airsim/home}
\end{abstract}

\begin{textblock*}{12cm}(5cm,1cm) % {block width} (coords) 
    {\setlength{\parindent}{0cm}
   \textcolor{red}{This work has been submitted to the IEEE International Conference on Robotics and Automation (ICRA) for possible publication. Copyright may be transferred without notice, after which this version may no longer be accessible}
   }
\end{textblock*}

\section{Introduction}

Visual Quadrotor navigation is the method of utilising vision sensors to navigate in an environment.  Very recently, researchers have utilised RL to perform various tasks such as: aerial filming \cite{goh_aerial_2021}, mapless exploration \cite{jang_hindsight_2022}, autonomous landing \cite{polvara_sim--real_2020}, counter drone operations \cite{cetin_counter_2021}, and collision avoidance \cite{shin_reward-driven_2020}.  However, RL can be very data inefficient, requiring a significant number of simulation steps for an agent to display favorable behaviours.  This is especially prevalent for flight simulators, where high fidelity environments and complex dynamic models lead to high computational costs.  The current issue within the field of visual quadrotor navigation is the time to train agents.  Currently training times can take tens for hours and sometimes even days.  Typically researchers will create toy problems for debugging purposes.  However, these are not always representative of the problem and there is no guaranty of an agent converging to a high episode reward.  Furthermore, RL fails silently which further emphasises the need for quicker training times.  Therefore, we present a simulation framework, built on AirSim, which provides efficient parallel training.  Furthermore, we incorporate a modified Ape-X networking architecture which can be used with AirSim's remote server to make use of decentralised training and numerous networked computers.  We show through experimentation the dramatic increase in efficiency of using 74 total agents with two computers and a speed up of training time from 3.9 hours to 11 minutes.

\section{Literature Review}

\subsection{Techniques to improve data inefficiency}

Improving the data efficiency of RL is a popular research area.  For roboticists, adding demonstration runs from an expert user or alternative controller can aid with better initialization.  Pienroj   use a PID ground\Sajad{-}truth controller at the start of the training run for powerline tracking \cite{pienroj_exploring_2019}.  Using an exploration training policy, the RL agent was able to refine the policy.  This approach presents difficulties; an expert might not be available, and it might be impractical to create an alternative controller for a given problem.

For off-policy based methods, a buffer is used to store transitions, which are sampled from, to train the neural network (NN).  Prioritising what samples are chosen can replay important transitions more frequently and hence learn more efficiently \cite{schaul_prioritized_2016}.

Multiple training environments and/or agents can be used to obtain observations in parallel.  Ape-X decouples acting from learning, where actors interact with separate instances of the environment and select actions according to a shared policy \cite{horgan_distributed_2018}.  This framework can be used with any off-policy algorithm such as Deep Q-Networks.  Additionally, off-policy agents can also be parallelised like such as Actor-Critic-based algorithms \cite{mnih_asynchronous_2016}.  Actor-Critic uses workers which send gradients, with respect to the parameters of the policy, to a central parameter server.  Similar to Ape-X, IMPALA is an alternative algorithm which communicates trajectories of experiences to a centralised learner \cite{espeholt_impala_2018}.  Based on the IMPALA architecture, asynchronous Proximal Policy Optimization (PPO) \cite{schulman_proximal_2017} is a policy gradient algorithm which uses a surrogate policy loss function with clipping.  More recently, Decentralised Distributed PPO provides further improvements by allowing each worker’s GPU to be used for both sampling and training \cite{wijmans_dd-ppo_2020}.

%Gradient updates on mini-batches of trajectories are sent from the actors to a server.  V-trace is an off-policy actor-critic algorithm  used to correct the lag between when actions are generated by actors and when the learner estimates the gradient.  Based on the IMPALA architecture, asynchronous Proximal Policy Optimization (PPO) \cite{schulman_proximal_2017} is a policy gradient algorithm which uses a surrogate policy loss function with clipping.  It is shown to be more efficient than synchronous PPO in wall-clock time due to the asynchronous sampling.  More recently, Decentralised Distributed PPO provides further improvements by allowing each worker’s GPU to be used for both sampling and training \cite{wijmans_dd-ppo_2020}.

Training distributed architectures in simulators, like the aforementioned, can benefit from physics acceleration.  Physics and frame rendering can take place asynchronously \cite{juliani_unity_2020}.  As a result it is possible to increase the speed of the simulation without increasing the frame rate of the renderer.  Simulator physics engines typically run on a fixed time-step, known as the clock speed.  Modifying this can change the speed of the simulator relative to the wall clock \cite{shah_airsim_2017}.  Faster simulators inherently lead to more observations gained and faster RL training times.  However, smaller time-steps can decrease the quality of the simulator.  Reduced quality simulations can lead to undetected collisions, which is not ideal for tasks like collision avoidance.  GPU-accelerated physics simulators have been shown to concurrently simulate hundreds of robots to train continuous-control locomotion tasks using Deep RL \cite{liang_gpu-accelerated_2018}.

\subsection{Robotic Simulators}

Advanced robotic simulators have been built on top of modern 3D games engines.  Self-driving cars benefit from Unreal Engine’s high fidelity renderer and physics engines.  For aerial vehicle simulators, AirSim \cite{shah_airsim_2017} also benefits from Unreal Engine but incorporate quadrotor dynamics for aerial vehicle navigation problems.  Flightmare \cite{song_flightmare_2021}, a more recent aerial simulator, is built on top of Unity’s platform.  They make use of parallel programming and decouple the quadrotor's dynamic modelling from the rendering engine for fast dynamic simulation.  Both quadrotor simulators make use of an asynchronous messaging library, Remote Procedural Call for AirSim and ZeroMQ for Flightmare.  Both libraries have used this architecture as it is currently not possible to train NNs using these game engines (Except for inference purposes using Onnx).  Other aerial simulators include RotorS \cite{koubaa_rotorsmodular_2016} using Gazebo, and Gym-pybullet-drones \cite{panerati_learning_2021} which have highly accurate physics engines but lack high fidelity rendering for visual-navigation tasks.  For a more detailed review on aerial simulators, the reader should direct their attention to this survey \cite{saunders_autonomous_2021}.  

\subsection{Reinforcement Learning for Robotic and quadrotor Navigation}

%Robotic
Early work of parallel deep RL had been used for visual-motor skill acquisition for physical robotic manipulators \cite{gu_deep_2016, yahya_collective_2017}.  More recently, using Isaac Sim, Ridin   trained a quadroped to walk with thousands of quadropeds in parallel which was validated on a real robot \cite{rudin_learning_2022}.  The authors used PPO with an observation space consisting of: velocities, joint position and velocities, previous actions, and distance measurements from the robot body to the terrain.  Finally, Song   \cite{song_autonomous_2021} uses Deep RL to compute near-time-optimal trajectories for drone racing which is tested on a physical quadrotor.  The state consists of relative gate observations and velocity, acceleration, orientation, and body rates.  They can simulate 100 environments in parallel but don't suffer from the rendering overhead of visual-navigation problems.

\subsection{Reinforcement Learning based Visual Quadrotor Navigation}

RL-based Visual quadrotor navigation techniques have been applied to a wide range of problems.  Goh uses RL for aerial filming \cite{goh_aerial_2021}.  Furthermore, RL has been explored for mapless exploration for quadrotrs.  Jang utilises hindsight intermediate targets within the environment and validates the approach on a ground robot \cite{jang_hindsight_2022}.  Autonomous landing is a well established field which has benefited from RL \cite{polvara_sim--real_2020}.  As technology on quadrotors progresses, security measures have become more prominent.  \c{C}etin uses RL to capture target quadrotors without crashing with obstacles within the environment \cite{cetin_counter_2021}.  Finally, collision avoidance has been thoroughly investigated.  Shin found continuous based Actor-Critic networks to perform best in cluttered environments \cite{shin_obstacle_2019}.  Furthermore, B\'ezier curves have been investigated as motion primitives for action selection \cite{camci_deep_2020}.  Generative Adversarial Networks have been explored to generate depth maps from a single image and then used as an input to an RL-based high-level planner to overcome computational power constraints on quadrotors \cite{singla_memory-based_2021}.   As can be seen in Table \ref{tab:papercomparison}, training times can take tens of hours and sometimes even days to complete.  More recently, researchers have utilised parallel architectures to improve training time.  Jang  train 5 soft Actor-Critic networks in parallel for mapless navigation  \cite{jang_hindsight_2022}.  It is unclear if 5 actors perform within the simulator or 5 individual instances of the unreal engine.  We take the phrasing of `5 worker threads' to convey the latter.  Running 5 individual instances of the Unreal Engine leads to huge overheads, instead we propose vectorising the environment such that these 5 agents can act within the same environment instance and benefit from the memory and processing efficiency.  This includes less assets being loaded in memory, allowing for larger replay buffers on centralised architectures and additional total agents for distributed architectures.  Similar to us, Fang   \cite{fang_quadrotor_2021} and Devo   \cite{devo_autonomous_2022} have used multiple agents within a single simulator instance.  Although the vectorised environment provides less computational burden compared to Jang's study, the authors could benefit from a decentralised architecture to utilise multiple environment instances.  

Our contributions are the following:     (1) Our study improves upon the state-of-the-art by providing a more effective way of vectorising the environment, increasing the number of agents within a single simulation instance.  To do so, we incorporate a non-interactive environment setup preventing rendering and collisions of all agents. Also, we improve the quantity of quadrotors in a given simulator by improving AirSim's functionality for parallel training using batched rendering and asynchronous episode learning which was not possible before.  (2) Finally we formulate a modified version of Ape-X, for AirSim simulations, to provide a decentralised learning framework to get up to $74$ quadrotors in two individual environments on two networked computers, although more computers can be used.

\section{Background}

\subsection{Algorithm Details}

Reinforcement learning (RL) is an agent-based modelling technique that studies the interactions between an environment.  For each time step $t$, an agent receives a state $s_t\in\mathcal{S}$ and selects an action $a_t\in\mathcal{A}$ according to the policy $\pi:\mathcal{A}\times\mathcal{S}\rightarrow[0,1],\quad \pi(a_t|s_t)$.  The state transition probability function $P$ captures the transition between the current state $s_t$ and the next state $s_{t+1}$.  Due to the Markov Property, the state captures information of past states meaning it is entirely independent of the past.  We can therefore express the transition probability as $Pr(R_{t+1}=r, S_{t+1}=s'|S_0, A_0, R_1, ...,S_{t-1}, A_{t-1}, R_{t}, S_t, A_t)=Pr(R_{t+1}=r, S_{t+1}=s'|S_t, A_t)$.  This Markov Decision Process can be defined as a tuple of these four components $(\mathcal{S}, \mathcal{A}, P, R)$.  The objective of the RL agent is to maximise the expected future cumulative discounted reward $V(s)=\mathbb{E}[G_t|S_t=s]$ and $G_t=\sum^T_{k=t+1}\gamma^{k-t-1}R_k$, where $\gamma$ is the discount factor $\gamma\in[0,1]$.  For deep Q-Learning, the aim is to estimate the state-action value function $Q(s,a)$ which can govern the agent's policy by calculating the expected returns for the state-action pairs.  In theory, if an agent can perform an infinite number of steps it would converge to the optimal action-value function $Q^*(s,a)$ using the Bellman equation as an iterative update.  Due to the impractical nature of infinite operations, functional approximators can be used in-place to estimate the action-value function.  For this study, we use a Deep Q-Network to which optimises the following function $Q(s,a)=\mathbb{E}_{(S,A,R,S')\sim\mathcal{D}}\left(R_{t+1}+\gamma\max_{a'}Q(s_{t+1}, a')|s_t=s, a_t=a\right)$, where $\mathcal{D}$ is the replay buffer.  DQN is an off-policy training algorithm, where the optimal policy is learnt independent of the agent's actions.  DQN utilises previous experiences by sampling from an experience replay buffer to perform the loss function.

\section{Method}

\subsection{Problem Formulation}

We showcase our parallel simulation platform by performing collision avoidance.  To do this, we utilise both an Inertial Measurement Unit (IMU) and depth camera which has been simulated within AirSim. The IMU provides linear velocity measurements, while the depth camera provides a mask of distances for each pixel within it's field of view.  We use Deep Q-Learning as a high-level controller \cite{mnih_human-level_2015} to learn the representation between the state and action to avoid obstacles.

\textbf{State:} The state space consists of depth images and linear velocity from IMU data, simulated within AirSim.  The Depth image has a resolution of $\left(32\times32\right)$ and the linear velocity has a resolution of $\left(1\times3\right)$.  We also apply image stacking, first introduced in Mnih's paper for Atari environments \cite{mnih_human-level_2015}, and now utilised for quadrotor visual navigation which incorporates memory into the state space.

\textbf{Action:}  We use a small action space consisting of left and right high level commands with a magnitude of $0.25\text{m/s}, \mathcal{A}=\{0.25, -0.25\}$.  The physics model time step is set to 4Hz and the action time step is 1s. 

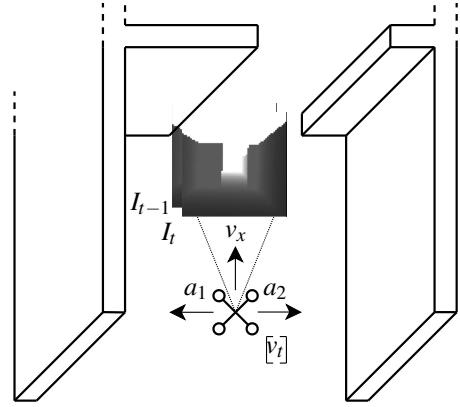
\begin{figure}[ht]
\centering
\begin{overpic}[width=6cm]{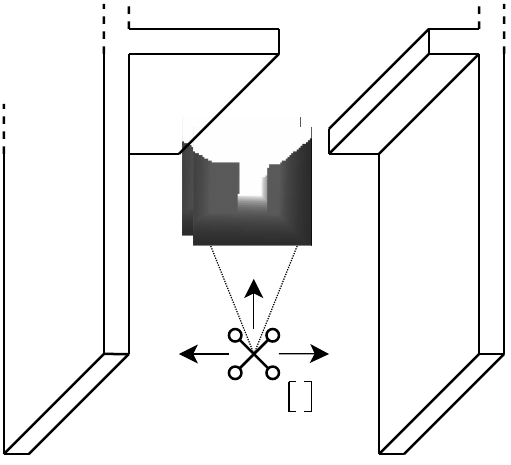}
\put(70,45){\makebox(0,0){$a_1$}}
\put(100,45){\makebox(0,0){$a_2$}}
\put(100,23){\makebox(0,0){$v_t$}}
\put(85,67){\makebox(0,0){$v_x$}}
\put(60,67){\makebox(0,0){$I_t$}}
\put(53,78){\makebox(0,0){$I_{t-1}$}}
\end{overpic}

\caption{{\small 2D visual navigation problem. In each time step, an observation is collected from the environment, consisting of the current depth image $I_t$ and the depth image collected at the previous time step $I_{t-1}$ with the current linear velocity $V_t$.  An action is chosen based on $\epsilon$-greedy action selection.  Each action is a high level command of either go left $a_1$ or go right $a_2$ which increments the current desired horizontal velocity for the lower level controller.  A constant forward velocity of $1m/s$ is also sent to the flight controller.}}
\label{fig:Pipeline}
\end{figure}

\begin{figure}[h]
\includegraphics[width=8cm]{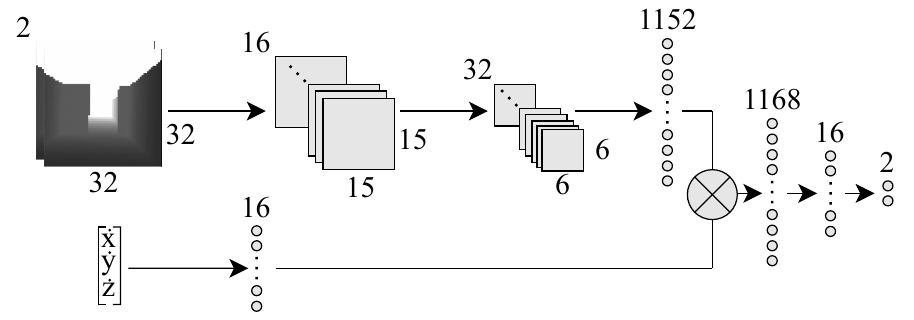}
\centering
\caption{{\small Deep Q-learn\Sajad{ing} network architecture, consisting of 2 inputs.  The depth image at time step $t$ and $t-1$ is fed into two layers of a convolutional layers.  The resultant latent space is flattened and concatenated with the output of the velocity terms which are fed through another fully connected network.  Finally, the concatenated layer is fed into two fully connected networks resulting in an output of 2 neurons representing the quality, Q value, of each action.}}
\label{fig:NetworkArchitecture}
\end{figure}

\textbf{Network Architecture:} The network architecture can be seen in Figure \ref{fig:NetworkArchitecture}.  The depth image is fed into a set of convolutional layers which incorporate the ReLu activation function and maxpooling.  The first convolutional layer has a kernel size of $\left(6\times6\right)$ and a stride of $\left(2\times2\right)$.  The second convolutional layer has a kernel size of $\left(3\times3\right)$ and a stride of $\left(1\times1\right)$.  The latent space from the convolutional layers are flattened into size $1152$.  The velocity of the quadrotor is fed into a fully connected layer and then concatenated with the flattened latent space.  Finally another two fully connected layers are used and the output vector size is the number of actions $|\mathcal{A}|$ which represents the quality values $Q(s,a)$.  These fully connected layers also use the ReLu activation function.  For the reinforcement learning agent, we use an experience replay size $|\mathcal{D}|$ of $15000$ and mini-batches of $32$.  The target network is updated every $150$ steps and a discount factor $\gamma$ of $0.99$ is used.  Before any training occurs, the experience replay is filled with $15000$ episodes.  Once filled, the Adam optimiser is used to update the parameter weights of the NN.  We linearly decrement $\epsilon$ using the number of experiences in the buffer $\epsilon=\max(0, 1-\frac{a_T}{|\mathcal{D}|})$.  Where $a_T$ is the total number of actions performed by all agents.

\textbf{Reward Function:} We use a simple reward function to train the agent.  A negative reward of $-100$ is given to the agent when it collides with the walls of the arena. Then, for each step it survives without colliding, the agent receives a positive reward of $+3$.

%\begin{equation}
%    R(s,a)= 
%\begin{cases}
%    -100,& \text{if collision}\\
%    3,              & \text{otherwise}
%\end{cases}
%\end{equation}

\subsection{Simulation Setup}

We modify AirSim \cite{shah_airsim_2017} to provide improved capabilities for parallel reinforcement learning.  We make use of the quadrotor physics model, sensor models and flight controller.  AirSim is a plugin for the Unreal Engine which provides a high fidelity rendering engine.  Using the Blueprint system we create an event graph which ignores overlapping events caused by other quadrotors, thus disabling agent collisions.  Furthermore we make all quadrotors hidden in the scene capture to prevent the depth map rendering other agents.  The benefit of these modifications is it allows us to vectorise the simulation and control multiple agents within the same Unreal Engine instance in a non-interactive way.  This leads to less complex environment setups with fewer required meshes to create individual areas for a quadrotor to navigate in.

\begin{figure}[h]
\centering
\begin{overpic}[width=8.5cm]{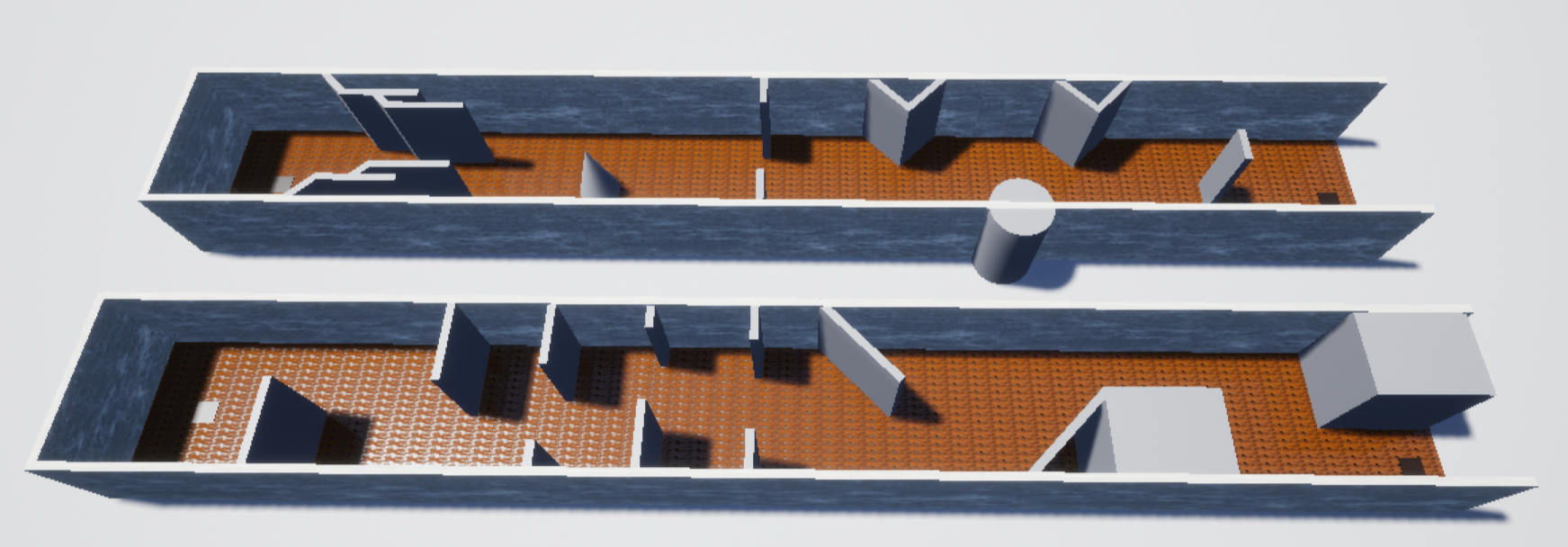}
\put(20,70){\makebox(0,0){(a)}}
\put(7,35){\makebox(0,0){(b)}}
\end{overpic}
\caption{{\small Environments to Train (b) and Test (a) the visual-based collision avoidance RL agent.  The aim is to navigate through the environment, starting on the left and traversing to the right without flying into a wall.}}
\label{fig:TrainTestArenas}
\end{figure}

The training and validation environments created within the Unreal Engine is depicted in Figure \ref{fig:TrainTestArenas}.  Both environments have a variety of obstacles which have been inspired from previous works of Camci \cite{camci_deep_2020} and Shin \cite{shin_reward-driven_2020}.  We understand recently, research has been done to apply RL collision avoidance controllers for very complex tasks.  Our aim here is illustrate a comparably competent controller through a more efficient training method.

\subsection{Synchronising Game and Render Thread}

The original command to request images from AirSim took the form {\tt\small simGetImages([ImageRequests], vehicle\_name, external)}.  The underlying issue with this command in the context of parallel reinforcement learning is the repetitive requests of individual images for each quadrotor.  An image request is an object which contains metadata on the image that is requested to be rendered.  This includes, for example, the type of image such as depth or RGB, resolution, and also FOV.  For this purpose, we require the same image metadata but for all vehicles within the simulator.  On-top of convenience, requesting multiple images this way can cause latency issues within the simulator.  AirSim requires a connection between the Unreal Engine and a python client using a Remote Procedure \Sajad{C}all (RPC) server.  This is due to incompatibility issues of using NN libraries within Unreal Engine.  The {\tt\small simGetImages} command is sent asynchronously and therefore, getting both physical properties of the camera and rendering the image at a specific timestep require synchronisation.  Hence, every call from {\tt\small simGetImages} causes a delay in order to process the request, which can scale with more agents in the simulator.

Therefore, we add the functionality to request multiple images from all vehicles within the simulator with one command in the same thread tick.  We call this new command {\tt\small simGetBatchImages([ImageRequests], [vehicle\_names])}.  We modify the AirSim plugin to incorporate the game logic to render all images within the same game thread.  To differentiate between both the origianl AirSim command  {\tt\small simGetImages} and our {\tt\small simGetBatchImages} command, we refer to these as non-batched and batched respectively.

%\begin{figure}[h]
%\includegraphics[width=8.5cm]{images/Method/BatchRenderGraph.pdf}
%\centering
%\caption{{\small Sketch illustrating thread synchronisation delay (orange) to render an image (green) using AirSim's {\tt\small simGetImages} command which is non-batched and our {\tt\small simGetBatchImages} command which obtains a batch of images within the same thread tick.}}
%\label{fig:BatchRenderGraph}
%\end{figure}

The limitation of using this command for all agents within the simulator is the synchronous image collection, requiring all agents to act simultaneously at the same timestep.  However, this limitation does not impact our training efficiency as we are training a single policy.

\subsection{Asynchronous Episodic Training}

Currently AirSim only contains the functionality to reset the entire simulator, placing all vehicles at the spawn location {\tt\small reset()}.  Once an agent has crashed, it waits until all other agents are finished before it can be reset.  This has an impact on the training time as experiences can only be gathered during flight.  Instead, we add the functionality to reset individual quadrotors and to spawn at a specified location with {\tt\small resetVehicle (vehicle\_name, pose)}.  This function provides the capability to train agents asynchronously and increase the rate at which experiences are generated. 

\subsection{Ape-X, Parallel Reinforcement Learning}

\begin{figure}[h]
\includegraphics[width=9cm]{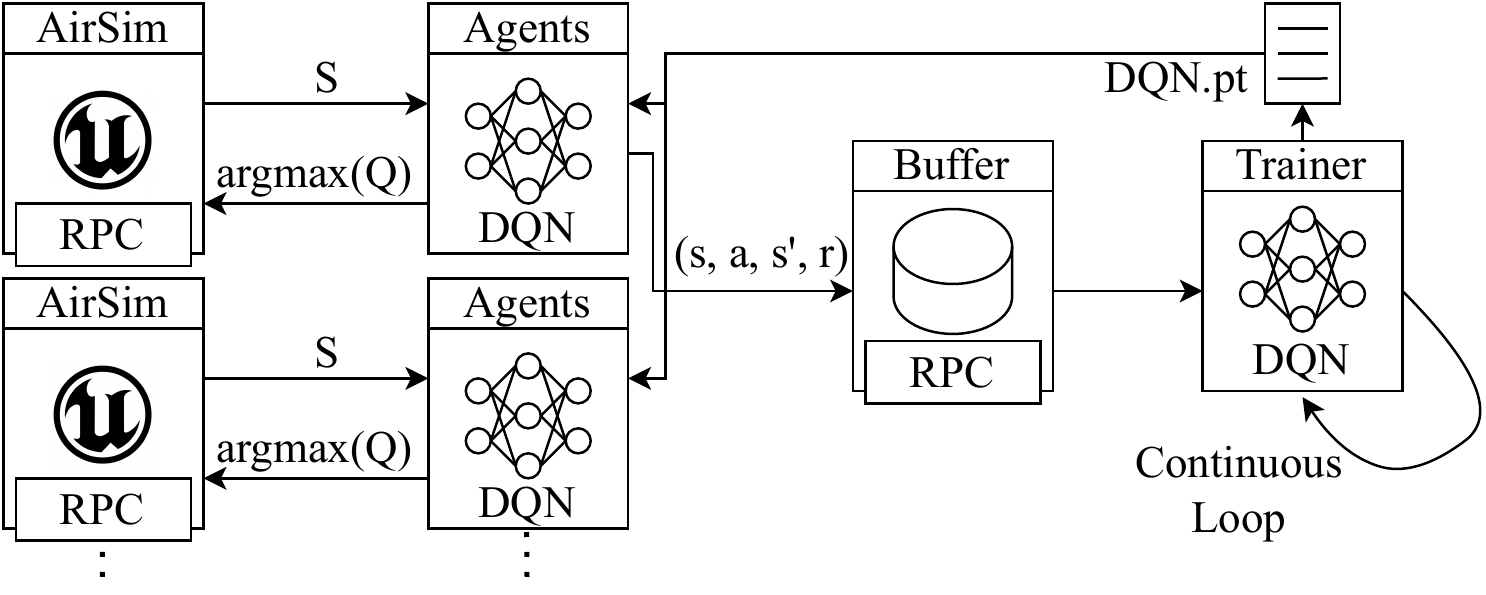}
\centering
\caption{{\small Decentralised networking architecture based off Ape-X consisting of local policy networks performing actions on AirSim instances.  Separate servers are used for the replay buffer and trainer which continuously samples from the buffer to train the global NN weights.}}
\label{fig:DistributedRL}
\end{figure}

We modify Ape-X to incorporate the additional RPC server node created within AirSim.  Figure \ref{fig:DistributedRL} illustrates the block diagram for the proposed decentralised network.  

%Local AirSim instances are connected to local instances of the neural network.  Experiences from the local instance are sent to a database server which is connected to the Trainer client.  This trainer client continuously samples mini-batches to perform gradient decent.  After every episode, local instances of the neural network are updated with the global neural network parameters from the Trainer client.

Local NN clients (Agents) are individually paired with AirSim instances.  For each local network, previous states are obtained using our batch render technique.  Actions are calculated using local instances of the global DQN network hosted on the Trainer client.  For every episode termination, the local NN parameters are updated with the global parameters from the Trainer client.  After every step of an episode, the experience consisting of the state, action, next state, and reward for each agent within the local AirSim instance, $(s, a, s', r)$ are sent to a database.  This database contains a set of the most recent experiences with first-in-first-out replacement.  Once the database is filled, the Trainer client continuously requests $32$ mini-batches of experiences to perform gradient descent and update the global NN parameters.  This framework is designed to be decentralised in different threads and communication via asynchronous messaging using RPC servers for AirSim instances and the experience replay database.  As a result, all blocks can run networked on different computers.  Performing gradient updates using a Trainer Client on a separate thread provides a major benefit.  The rate of gradient calculations is not synchronous to the action time such as in \cite{kersandt_self-training_2018}.  For all experiments, we keep the training frequency constant at $50$Hz.

\begin{table*}[bp!]
\caption{Comparison of state-of-the-art work, illustrating the training time for various problems and the state space used.  Furthermore, computer specification is listed and the number of agents used to train the policy.  Our work increases the number of agents, reduces training time using similar hardware than other works.}
\centering
\begin{tabularx}{1\textwidth}{ | p{2.5cm} | p{0.75cm} |  p{0.75cm} |  p{0.75cm} | p{1.3cm} | p{1cm} | p{1cm} |  p{2cm} | X | p{0.75cm} | }
\hline\hline
\multirow[c]{2}{2cm}{\textbf{Author}} &
\multicolumn{3}{c|}{\textbf{Agents}} &
\multirow[c]{2}{2cm}{\textbf{Image State Dimensions}} &
\multicolumn{2}{c|}{\textbf{Training}} &
\multicolumn{3}{c|}{\textbf{Compute Power}} \\
\cline{2-4}
\cline{6-10}
 &
Envs &
Agents &
Total &
 &
Steps &
Time &
CPU &
GPU &
RAM \\
\hline

Kersandt 2018 \cite{kersandt_self-training_2018} &
1 &
1 &
1 &
$\left(30\times100\right)$ &
150,000 &
41 hours &
- &
- &
- \\

\hline

Shin 2019 \cite{shin_obstacle_2019} &
1 &
1 &
1 &
$\left(4\times64\times64\right)$ &
100,000 &
- &
Intel i7 3.4 GHz &
Nvidia Titan X &
- \\
 
 \hline

\c{C}etin 2020 \cite{cetin_counter_2020} &
1 &
1 &
1 &
$\left(30\times100\right)$ &
50,000 &
14 hours &
Intel i7 &
Nvidia GTX 1060 &
16GB \\

\hline

Polvara 2020 \cite{polvara_sim--real_2020} &
1 &
1 &
1 &
$\left(4\times84\times84\right)$ &
- &
7.6 days &
Intel i7 &
Nvidia Tesla K-40 &
32GB \\
 
 \hline
 
He 2020 \cite{he_deep_2020} &
1 &
1 &
1 &
$\left(80\times100\right)$ &
50,000 &
- &
- &
- &
- \\

\hline

\c{C}etin 2021 \cite{cetin_counter_2021} &
1 &
1 &
1 &
$\left(30\times100\right)$ &
75,000 &
48 hours &
Intel i7 &
Nvidia GTX 1060 &
16GB \\

\hline
 
Singla 2021 \cite{singla_memory-based_2021} &
1 &
1 &
1 &
$\left(84\times84\right)$ &
300,000 &
- &
Intel i7 &
Nvidia GTX 1050 &
8GB \\
 
 \hline
 
Fang 2021 \cite{fang_quadrotor_2021} &
1 &
9 &
9 &
$\left(64\times64\right)$ &
- &
29 mins &
Intel i7-9700 &
Nvidia RTX 2080Ti &
64GB \\
 
 \hline
 
Jang 2022 \cite{jang_hindsight_2022} &
5 &
1 &
5 &
$\left(3\times84\times84\right)$ &
4,000,000 &
33h &
Intel i9-9900K &
Nvidia RTX 2080Ti &
- \\
 
 \hline

Devo 2022 \cite{devo_autonomous_2022} &
1 &
8 &
8 &
$\left(84\times84\right)$ &
2,700,000 &
- &
Intel i9-9900K &
2×Nvidia RTX 2080Ti &
64GB \\
 
 \hline
 
\textbf{Ours} &
2 &
37 &
74 &
$\left(2\times32\times32\right)$ &
65,000 &
11 mins &
Intel i7-11700k &
Nvidia RTX 2070 &
64GB \\
\hline\hline

\end{tabularx}
\label{tab:papercomparison}
\end{table*}

\section{Results}

First, the batched and non-batched synchronisation techniques were compared. A sweep of different agent configurations is conducted which analyised the time to obtain the quadrotor state.  Then we further illustrate the cumulative delay caused by continuing synchronisation of the non-batched method compared to our batched method.  Finally, we show how efficient our parallel framework is for the given collision avoidance task.  Comparing the training time against the number of agents within the simulator.

\subsection{Batched and Non-Batched Comparison Results}
\begin{figure}[h]
\centering
\begin{overpic}[width=8.5cm]{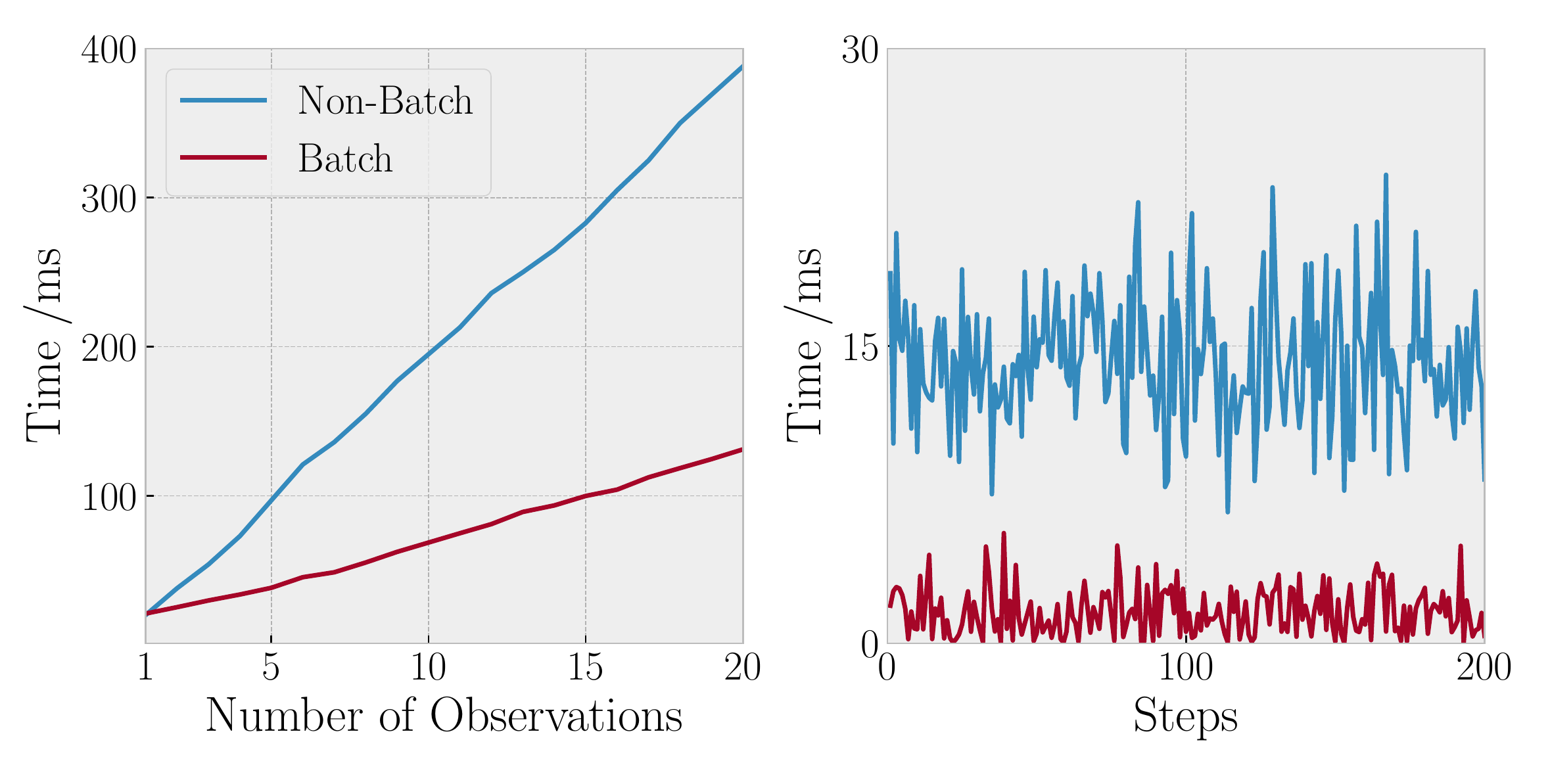}
\put(15,15){\makebox(0,0){(a)}}
\put(130,15){\makebox(0,0){(b)}}
\end{overpic}
  \caption{{\small (a) presents the computation time to render quadrotor states for both batched and non-batched techniques.  (b) presents the magnitude of the delay caused by synchronising the game and render thread either once or multiple times for $10$ quadrotors, using the batched and non-batched method.}} \label{fig:synchronisingResults}
\end{figure}

First, we analyse the computational time to render the quadrotors state for both the batched and non-batched techniques.  Figure \ref{fig:synchronisingResults}(a) shows the total average time, for each timestep, to calculate all quadrotor states within the simulator.  As illustrated, our method renders quadrotor states quicker than the non-batched variant.  Calculating $10$ quadrotor states takes $197$ms for the non-batched technique, whereas our method can render the same states in $61$ms, more than half the time.

Further analysis, in Figure \ref{fig:synchronisingResults}(b), shows the magnitude of the delay caused by repeatedly synchronising the game and render threads.  For $10$ quadrotors, we calculated the synchronisation delay to be an average of $14.2$ms.  Whereas, our method is only synchronising once per request, hence an average delay of $1.6$ms is recorded.  This result does not explain the entire improved efficiency and we believe further optimisations occur within the Unreal Engine.

\subsection{Parallel Collision Avoidance Agents}

\begin{figure}[h]
\centering
\begin{overpic}[width=8.5cm]{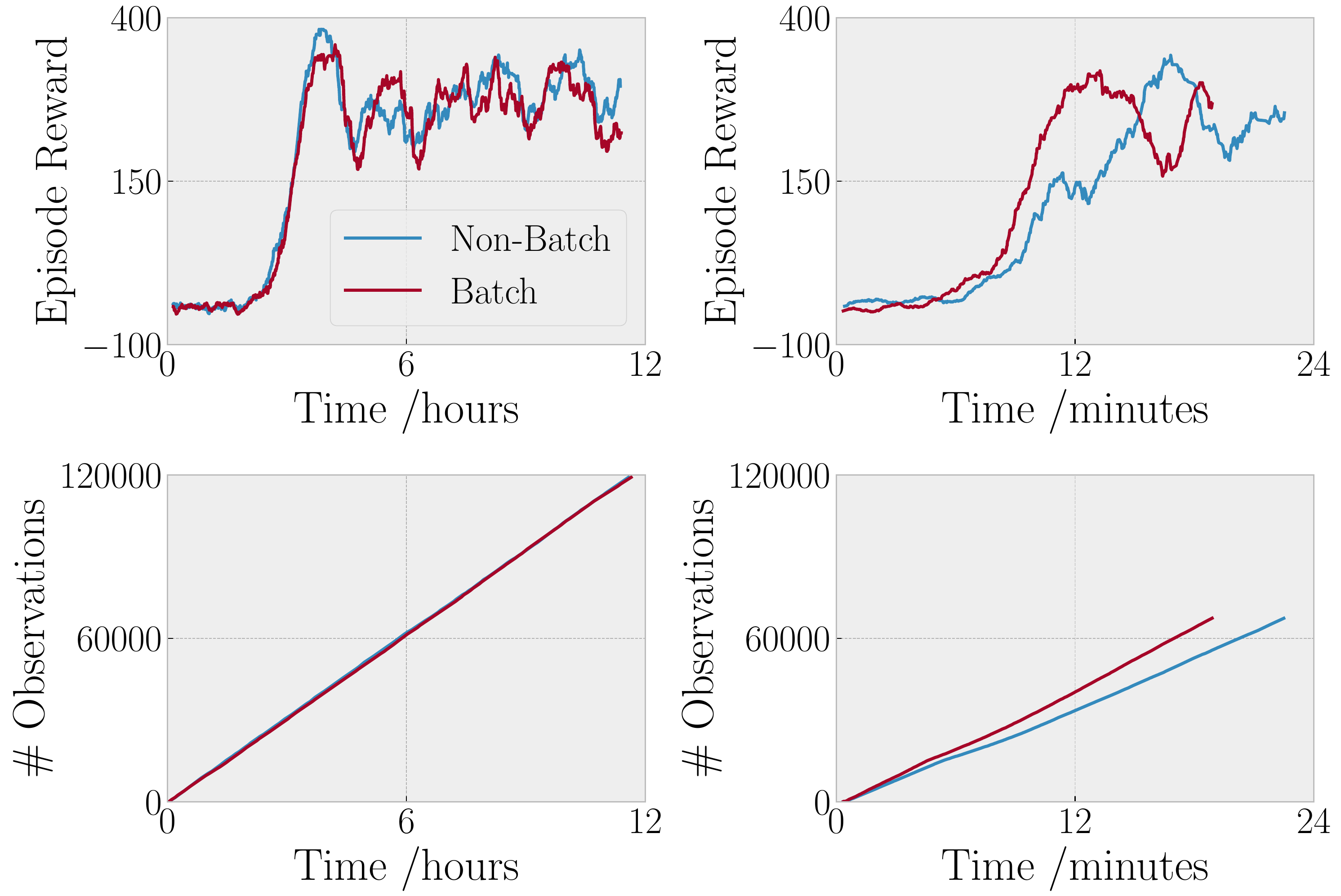}
\put(20,90){\makebox(0,0){(a)}}
\put(20,10){\makebox(0,0){(b)}}
\put(140,90){\makebox(0,0){(c)}}
\put(140,10){\makebox(0,0){(d)}}
\end{overpic}
  \caption{{\small Comparison of episode reward and number of observations rendered for both batched and non-batched methods using $1$ (a),(b) and $50$ (c),(d) quadrotors in parallel.  Reduction in training time is observed for $50$ quadrotors, but not for $1$ quadrotor.}} 
  \label{fig:parallelRewardObservations1and50}
\end{figure}
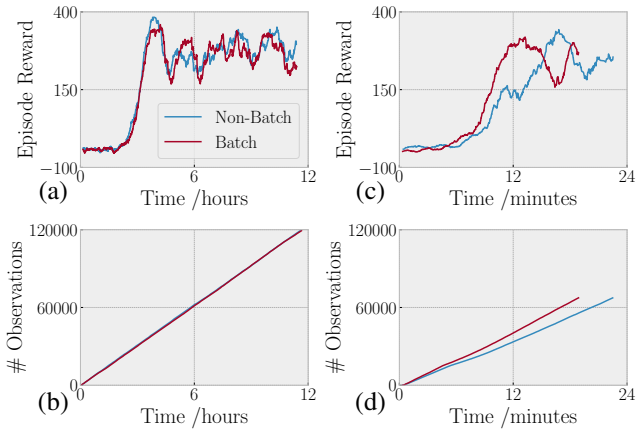

Using the new render technique, we show the efficiency of using Ape-X with a fully vectorised environment.  The vectorised environment involves multiple non-interactive agents and multiple environments all running in parallel to train a single policy.  We present a comparison, shown in Figure \ref{fig:parallelRewardObservations1and50},  of  both the batched and non-batched methods for both a single quadrotors and $50$ quadrotors running in parallel.  The rate at which states were rendered for both methods of a single quadrotor was similar, as expected.  This is because the thread is only being paused once for both methods and as a result observation sample rate was the same, shown in Figure \ref{fig:parallelRewardObservations1and50}(b).  Whereas for $50$ quadrotors, obtaining observations in batches resulted in quicker training time as a result of the increased rendering efficiency, shown in Figure \ref{fig:parallelRewardObservations1and50}(c).

All experiments are tabulated in Table \ref{tab:trainingrun}.  Tests were performed using an Intel i7-11700K CPU, Nvidia RTX 2070 Super GPU, and 64GB of RAM.  To illustrate the computational power used when training, metrics on CPU and GPU utilisation with GPU power consumption were collected.  Training time is the time for the model to converge to an episode reward of 300.  We validated our model on the test arena, shown in Figure \ref{fig:TrainTestArenas}, for each experiment.  On average the success rate was 78\%, showing good generalisation to new environments. 

%Training time
As expected, increasing the number of agents collecting experiences resulted in dramatically shorter training time.  Furthermore, we noticed a configuration of $25$ quadrotors acting in parallel using batched rendering achieved roughly the same time to converge as the non-batched method.  The fastest training time was achieved with $74$ agents acting in parallel over $2$ environments with $11$ minutes to converge.  Whereas, the slowest total training time took $3.9$ hours to converge.

%RAM
The total RAM utilised by the distributed framework for a single environment amounts to $7.3$GB.  Furthermore, $12.6$GB of RAM was used for two simulation instances of AirSim but split between two independent computers.  Although, we only utilised approximately $2$GB of RAM for the replay buffer, larger state spaces for more complicated tasks such as the ones identified in Table \ref{tab:papercomparison} illustrate the need for additional RAM.  Distributing the computational resources over many computer allows for larger state spaces and replay buffer sizes.

\begin{table}[h]
\caption{{\small Comparison between the parallel architecture and rendering technique.  Batched rendering performed slightly better, however CPU utilisation sharply rose at 50 agents.  This was solved by using two instances of AirSim to train a total of 74 agents.}}
\centering
\begin{tabularx}{0.5031\textwidth}{ | p{1.5cm} | p{0.5cm} |  p{0.75cm} |  p{0.5cm} | p{0.5cm} |  p{0.5cm} | p{0.5cm} | p{0.6cm} |}
\hline\hline
\multirow[c]{2}{2cm}{\textbf{Threading Method}} &
\multicolumn{3}{c|}{\textbf{Agents}} &
\multirow[c]{2}{0.5cm}{\textbf{Train Time}} &
\multicolumn{2}{c|}{\textbf{Utilisation\%}} &
\multirow[c]{2}{1cm}{\textbf{GPU Power}} \\
\cline{2-4}
\cline{6-7}

 &
Envs &
Agents &
Total &
 &
CPU &
GPU &
 \\
\hline

\multirow[c]{2}{2cm}{Non-Batched Observations} &
1 &
1 &
1 &
3.8h &
24\% &
53\% &
101W \\

  &
1 &
50 &
50 &
18m &
62\% &
59\% &
95W \\
\hline

\multirow[c]{4}{2cm}{Batched Observations} &
1 &
1 &
1 &
3.9h &
24\% &
54\% &
92W \\

  &
1 &
25 &
25 &
17m &
27\% &
53\% &
93W \\

  &
1 &
50 &
50 &
14m &
73\% &
55\% &
95W \\

  &
2$^*$ &
37 &
74 &
11m &
31\% &
56\% &
91W \\
\hline\hline
\multicolumn{8}{l}{$^*$ Environment instances run on networked computers to overcome the} \\
\multicolumn{8}{l}{CPU bottleneck.} \\

\end{tabularx}

\label{tab:trainingrun}
\end{table}

%CPU
We notice CPU usage dramatically increases after $37$ quadrotors.  Therefore, it can be assumed for any given computer specification, a sweet spot exists where a maximum number of quadrotors can be used for the fastest throughput.  For this, we would need to run further experiments.  We attempted to run simulations with $100$ and $200$ quadrotors in a single environment with no increase in performance.  For our setup, the sweet-spot was approximately $50$ quad rotors per environment as shown by the $73\%$ CPU utilisation.

%GPU
GPU utilisation and power were not dependent on the number of quadrotors.  This leads us to believe, alongside the results of the CPU utilisation, that the bottleneck for number of quadrotors per environment lies with the CPU performance.   

\begin{figure}[h]
    \centering
    \begin{overpic}[width=8.5cm]{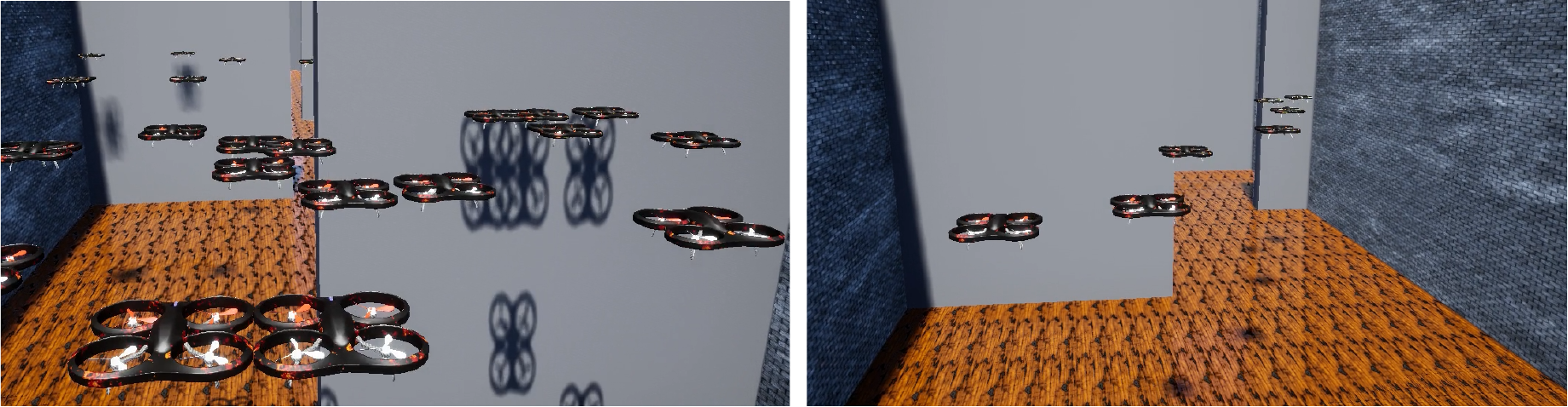}
    \put(60,-5){\makebox(0,0){(a)}}
    \put(185,-5){\makebox(0,0){(b)}}
    \end{overpic}
    \vspace{1mm}
    \caption{{\small Comparison between the exploration (a) and exploitation (b) behaviour during training.}} 
    \label{fig:TrainedUntrainedDrones}
\end{figure}

Figure \ref{fig:TrainedUntrainedDrones} illustrates the behaviour of the quadrotors during training.  As the agents start to collect experiences, the $\epsilon$-greedy action selection results in initially random actions in an attempt to explore the environment (a).  As soon as the experience replay buffer is filled, the $\epsilon$ value decreases and the agent starts to exploit the policy (b).

%Training frequency
We also noticed the frequency of gradient calculations of the NN droped when multiple AirSim instances were run on a single computer.  For two instances, the training frequency resulted in $14$Hz, further illustrating the need for the decentralised parallel framework we present here.

\section{Conclusion}

%Results
In this paper, we improve upon the state-of-the-art by providing a more efficient way of parallel agent training.  To do so, we incorporate a non-interactive environment setup, which prevents rendering and collisions of all agents. Using a non-interactive vectorised setup also leads to fewer required meshes to create individual areas, improving the computational performance.  We incorporated a batch rendering method, which prevented redundant game and render synchronisation which lead to increased throughput and reduced training time.  With our modified Ape-x architecture, and asynchrounous episodic training, we were able to run $74$ quadrotors in parallel in two individual environments on two network computers, 10 fold greater than current state-of-the-art.  The increased agent count dramatically reduced training time down to $11$ minutes from $3.9$ hours.  This technique allows greater utilisation of RAM for larger replay buffer requirements.  Furthermore, we show the that CPU resource utilisation is the bottleneck in our system.  We overcome this by distributing the computational resources over many networked computers.  Future work will investigate how the performance of the CPU can be improved using techniques such as GPU-accelerated physics engines.  Furthermore, we will test an agent trained using our framework on a physical quadrotor, investigating the performance of sim-to-reality gap.

\bibliographystyle{ieeetr}
\bibliography{references.bib}

\end{document}